\documentclass[letterpaper, 10 pt, conference]{ieeeconf}  

\IEEEoverridecommandlockouts                              

\usepackage{graphics} 
\usepackage{epsfig} 
\usepackage{mathptmx} 
\usepackage{times} 
\usepackage{amsmath} 
\usepackage{amssymb}  
\usepackage{booktabs}
\usepackage{array}
\usepackage{graphicx}
\usepackage{float}
\usepackage{subfigure}
\usepackage{multirow}
\usepackage{threeparttable}
\usepackage{hyperref}

\title{\LARGE \bf
Efficient Camera Exposure Control for Visual Odometry via Deep Reinforcement Learning
}

\author{Shuyang Zhang$^{1}$, Jinhao He$^{2}$, Yilong Zhu$^{1}$, Jin Wu$^{1}$ and Jie Yuan$^{1}$
\thanks{$^{1}$S. Zhang, Y. Zhu, J. Wu and J. Yuan are with the Department of Electronic and Computer Engineering, the Hong Kong University of Science and Technology, Clear Water Bay, Kowloon, Hong Kong SAR, China. (email: szhangcy@connect.ust.hk)}
\thanks{$^{2}$J. He is with the Thrust of Robotics and Autonomous Systems, the Hong Kong University of Science and Technology (GZ), Guangzhou, China.}
}

\begin{document}

\maketitle
\thispagestyle{empty}
\pagestyle{empty}

\begin{abstract}
    The stability of visual odometry~(VO) systems is undermined by degraded image quality, 
    especially in environments with significant illumination changes. 
    This study employs a deep reinforcement learning~(DRL) framework to train agents for exposure control, 
    aiming to enhance imaging performance in challenging conditions.
    A lightweight image simulator is developed to facilitate the training process, 
    enabling the diversification of image exposure and sequence trajectory.
    This setup enables completely offline training, eliminating the need for direct interaction with camera hardware and the real environments.
    Different levels of reward functions are crafted to enhance the VO systems, 
    equipping the DRL agents with varying intelligence.
    Extensive experiments have shown that our exposure control agents achieve superior efficiency—with an average inference duration of 1.58 ms per frame on a CPU—and respond more quickly than traditional feedback control schemes. 
    By choosing an appropriate reward function, agents acquire an intelligent understanding of motion trends 
    and can anticipate future changes in illumination. 
    This predictive capability allows VO systems to deliver more stable and precise odometry results. 
    The code and dataset are open source on \url{https://github.com/ShuyangUni/drl_exposure_ctrl}.
\end{abstract}

\section{Introduction}

Effective camera exposure control is crucial for dynamic robotics applications like visual odometry~(VO), 
characterized by complex lighting. 
Inadequate exposure control may fail to promptly adjust to the rapid changes in lighting or field of view, 
resulting in protracted periods of over-saturation (under-exposure or over-exposure) 
and consequent loss of critical information, which poses safety risks to the robots.

Traditional exposure control methods are divided into gradient-based~\cite{shim2014auto,zhang2017active,han2023camera} and function-fitting approches~\cite{kim2020proactive, zhang2024image}, 
each can attain satisfactory outcomes under standard conditions. 
Nevertheless, these methods manifest limitations in environments with drastical illumination changes, 
such as swift transitions through tunnels~\cite{tomasi2021learned}.
The gradient-based methods suffer from restricted feedback control speed, 
resulting in the inability to update exposure parameters with rapid scene alterations, 
consequently leading to overexposure or underexposure. 
The function-fitting approaches, formulating a correlation between image metrics and exposure, 
provide an instantaneous, one-shot solution adaptable to various metric designs. 
However, the substantial computational demand leads to infrequent parameter updates. 
While this frequency limitation may prevent underexposure, 
abrupt parameter adjustments can introduce significant variability in image content, 
thereby undermining feature tracking efficacy.

\begin{figure}[t]
    \centering
    \includegraphics[width=\linewidth]{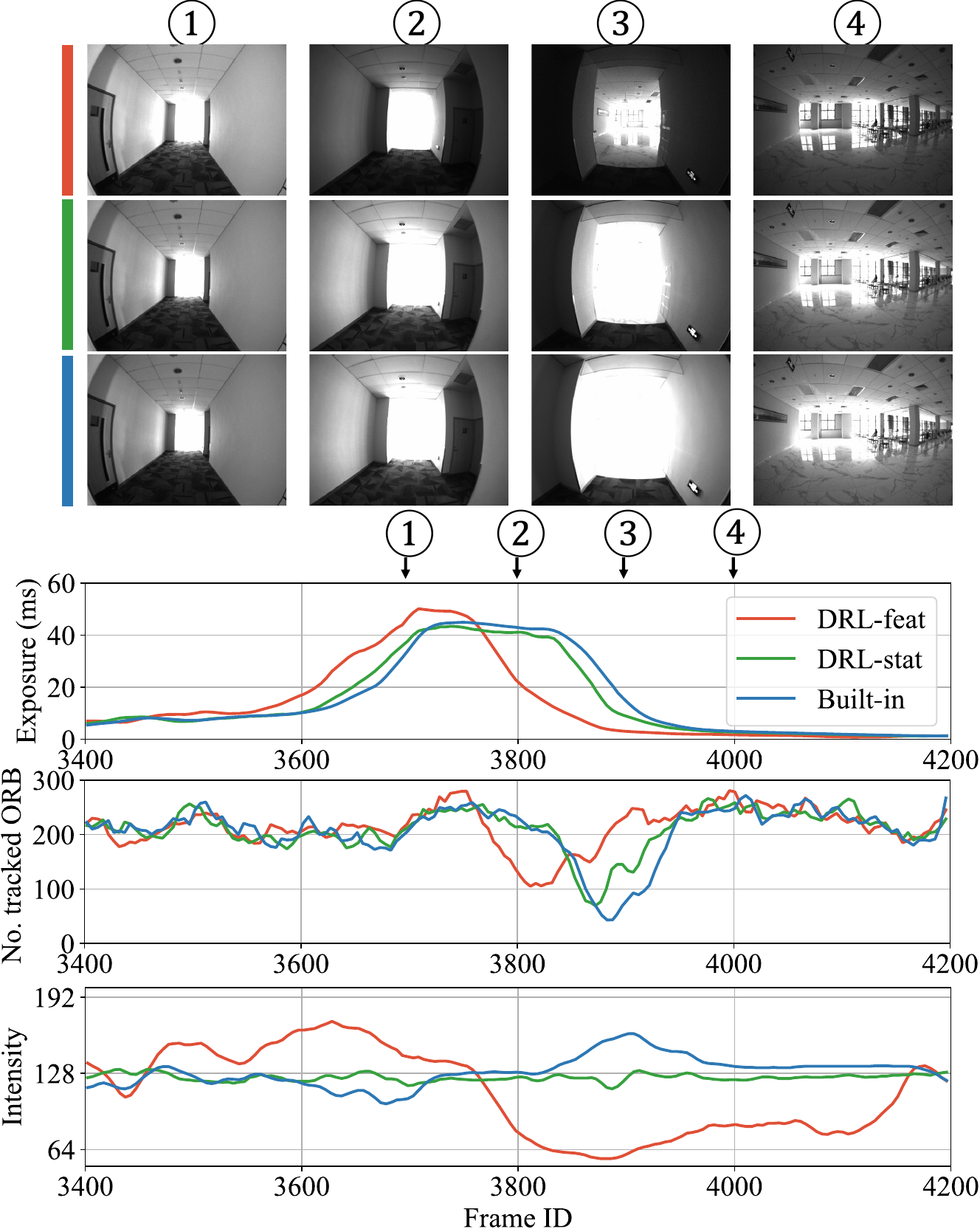}
    \caption{An illustration of drastic illumination change in our \emph{Corridor} sequence.
        Our DRL-based method with feature-level rewards~(\textbf{DRL-feat}) 
        exhibits a high-level comprehension of lighting and motion, 
        surpassing the traditional method~(\textbf{Built-in}) 
        and DRL method with image statistic-level rewards~(\textbf{DRL-stat}).
        The agent \textbf{DRL-feat} predicts the impending over-exposure event and preemptively reduces the exposure. 
        While this adjustment temporarily decreases the number of tracked feature points, 
        it effectively prevents a more severe failure in subsequent frames.
    }\label{fig_diff_cover}
\end{figure}

Several researchers~\cite{tomasi2021learned,lee2024learning} have incorporated 
deep reinforcement learning~(DRL) into exposure control tasks, 
yielding numerous benefits. 
DRL-based systems employ an end-to-end framework from image inputs to control output, 
obviating the need for complex system design. 
the reward-based nature of reinforcement learning facilitates the integration of direct optimization targets, 
such as the number of features. 
These targets, which are non-differentiable and labor-intensive to compute using traditional techniques, 
are readily accommodated within the DRL framework.
However, current DRL-based methods that rely on online interaction face multiple challenges. 
The design of the requisite custom hardware is complex 
and is difficult for the comunity to reimplement. 
Additionally, the mode of online interaction leads to inefficient data sampling, 
where each data sequence is employed only once and becomes obsolete after a policy update. 
This inefficiency in data utilization hampers the system's ability to adequately update the policy within the constraints of limited training periods. 
Therefore, to enhance training stability, it becomes necessary to integrate specific mechanisms~\cite{tomasi2021learned} or employ curriculum learning~\cite{lee2024learning}.

In this study, we also propose a DRL-based exposure control method, 
focusing on the intelligence and robustness of the VO system under complex lighting conditions. 
We decouple data collection from model training by adopting an offline training approach to improve sample efficiency. 
Compared to prior methods, our framework demonstrates enhanced convergence efficiency and simplicity.
The contributions of our work include:
\begin{itemize}
    \item A DRL-based camera exposure control solution.
          The exposure control challenge is divided into two subtasks, 
          enabling completely offline DRL operations without the necessity for online interactions.
    \item An lightweight image simulator based on imaging principles,
          significantly enhances the data efficiency and simplifies the complexity 
          of DRL training.
    \item A study on reward function design with various levels of information.
          The trained agents are equipped with different intelligence, 
          enabling them to deliver exceptional performance in challenging scenarios.
    \item Sufficient experimental evaluation, which demonstrates 
          that our exposure control method improves the performance of VO tasks, 
          and achieves faster response speed and reduced time consumption.
\end{itemize}

\section{Related Work}

\subsection{Optimization-based Exposure Control}
Traditional exposure control methods are commonly conceptualized as optimization problems, 
addressed through either \emph{gradient-based} or \emph{function-fitting} techniques.

Gradient-based methods~\cite{shim2014auto, zhang2017active, han2023camera}
compute the gradient of the image metric relative to exposure 
to determine the direction for parameter updates. 
These methods employ iterative optimization to refine exposure settings progressively. 
Shim~\textit{et al.}~\cite{shim2014auto}
applied $\gamma$-correction to simulate exposure variations,
assessing image gradient magnitudes.
Zhang~\textit{et al.}~\cite{zhang2017active} enhanced Shim's metric 
and rigorously validated their derivatives.
Han~\textit{et al.}~\cite{han2023camera} integrated optical flow to evaluate camera self-motion, 
thus enabling adjustments to gain and exposure time to mitigate motion blur.
These methods typically exhibit \emph{reactive} behavior, 
adjusting only in response to changes in scene conditions, 
which can lead to over-saturation in dynamically lit environments.

Function-fitting methods~\cite{kim2020proactive, zhang2024image}
utilize synthetic imaging techniques to generate images at varied exposure levels, 
targeting global optimality directly. 
Kim~\textit{et al.}~\cite{kim2020proactive}
developed a mixed image quality metric 
and produced synthetic images from a fixed exposure seed image, 
with Bayesian optimization to simultaneously refine exposure time and gain settings. 
Zhang~\textit{et al.}~\cite{zhang2024image} integrated image bracketing patterns 
with synthetic techniques to periodically survey a high dynamic range~(HDR) spectrum of the scene.
Function-fitting approaches provide rapid responses,
potentially offering optimal settings from a single image capture; 
however, they require significant computational resources due to the exhaustive global optimization process.

\subsection{Reinforcement Learning-based Exposure Control}
Recent advancements in deep reinforcement learning~(DRL) has markedly broadened 
their applications in real-world robotic tasks, 
including locomotion control~\cite{romero2023actor, cai2022dqgat}, 
perception~\cite{jeong2021deep, messikommer2024reinforcement}, 
and decision making~\cite{cheng2024pluto, xing2024bootstrapping}.

Existing DRL-based exposure control methods~\cite{tomasi2021learned, lee2024learning} suffer the \textbf{online} interative training scheme.
Tomasi~\textit{et al.}~\cite{tomasi2021learned} utilized a convolutional neural network~(CNN) 
to dynamically adjust camera gain and exposure time settings. 
They devised a self-supervised labeling technique 
using dual cameras and a specialized capture pattern. 
However, their approach of online data collection combined with offline agent training 
presents complexities and inefficiencies in implementation.
Additionally, the infrequent policy updates within their system 
may hinder the generalization capabilities of the agent.
Lee~\textit{et al.}~\cite{lee2024learning} proposed a device-specific method to facilitate online training. 
They vectorized input images to eliminate CNN modules, thereby reducing model complexity. 
They introduced a progressive lighting curriculum 
that incrementally exposed the agent to increasingly complex training scenarios.
Nonetheless, the dependence on a specialized stable device limits their method's generalizability and fails to address camera motion.
Moreover, their approach to image vectorization sacrifices essential high-level visual information, 
such as feature detection and tracking.

In this study, 
we develop a DRL-based camera exposure control method 
specifically designed for VO systems.
Our approach utilizes cameras equipped with a Sequencer mode, 
a bracketing imaging pattern that captures images at varying exposures, 
a feature prevalent in advanced machine vision cameras 
such as the FLIR Blackfly S\footnote{https://www.flir.com/products/blackfly-s-usb3/} and Basler ACE\footnote{https://www.baslerweb.com/en/cameras/ace/}.
We leverage this bracketing pattern alongside image synthesis techniques 
to construct an environment simulator for training exposure control agents. 
By introducing an intermediary exposure parameter, 
we dissect the exposure control problem into a cascading dual-module pipeline.
The first module determines the optimal exposure, 
facilitating entirely \textbf{offline} training via our simulator. 
The second module implements a rule-based strategy to allocate the optimal exposure 
to exposure time and gain, interfacing directly with camera hardware.
The environment simulator enhances the reward function by integrating multiple information tiers, 
thereby enabling agents to exhibit varied levels of intelligence within the VO systems.

\begin{figure*}[t]
    \centering
    \subfigure[Training process.\label{fig_fw_train}]{
        \centering
        \includegraphics[width=0.95\linewidth]{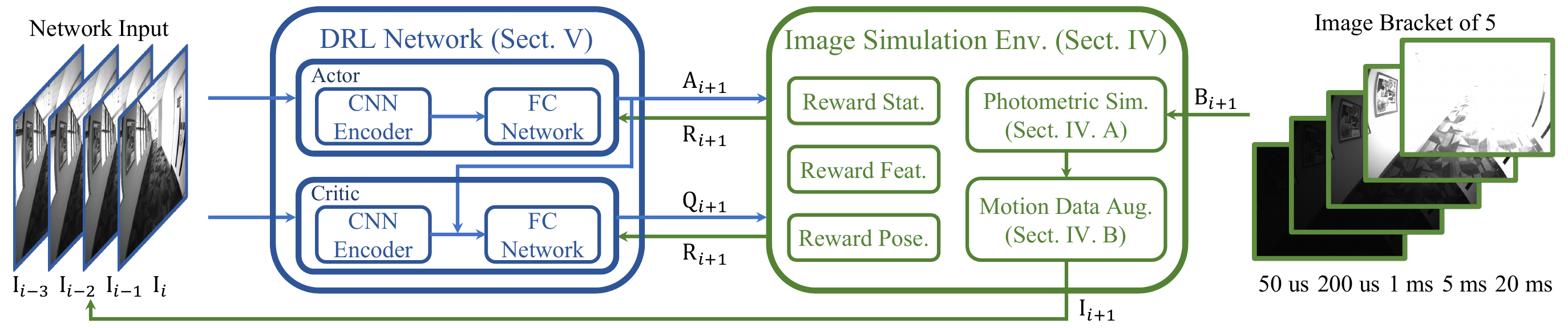}} \\ \vspace{-0.35cm}
    \subfigure[Inference process.\label{fig_fw_infer}]{
        \centering
        \includegraphics[width=0.95\linewidth]{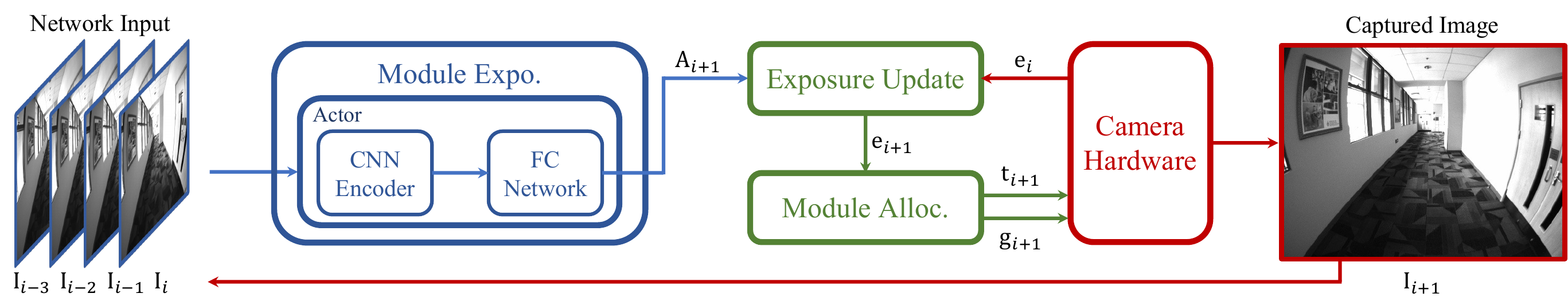}}
    \caption{System overview of the training and inference phases.
    In the training phase, we employ the image bracketing technique for simulation, enhanced by data augmentation to diversify sequence motion.
    The Soft Actor-Critic (SAC) framework was adopted for our DRL implementation. 
    Within this framework, agents (actors) of varing intelligence levels were trained using distinct reward designs.
    In the inference phase, these trained agents generate a continuous relative action signal, 
    which is then translated into target exposure for the next image. 
    These control signals, comprising both exposure time and analog gain, 
    are allocated via a rule-based strategy before transmission to the camera hardware. 
    The newly captured image is subsequently fed back into the agent's input for ongoing inference.
    }\label{fig_fw}
\end{figure*}

\section{System Overview}
\subsection{Traditional Problem Formulation}
Camera exposure control can be conceptualized as an optimization problem, 
formally expressed as
\begin{equation}
    e_{i+1} = \mathop{\arg \max}_{e} \mathbf{M}(\mathbf{I}(e_i)),
\end{equation}
where $e$ denotes the \emph{exposure}, 
a composite variable defined by exposure time $t$ and analog gain $g$
\begin{equation}
    e = t \times 10^{\frac{g}{20}}.
\end{equation}
Upon determining the optimal exposure $e$, 
an attribute allocation decomposes the exposure back into its constituent parameters, 
exposure time and analog gain, which are then used to adjust the camera's hardware settings directly.
The function $\mathbf{I}(\cdot)$ denotes the imaging process 
that yields an image $\mathbf{I}_i$ for a specified exposure,
while $\mathbf{M}(\cdot)$
assesses an image according to pre-defined metrics, 
such as gradient magnitude~\cite{shim2014auto, zhang2017active}, 
entropy~\cite{han2023camera, kim2020proactive} and their combinations.

To address this optimization, 
feedback control methods compute an update increment $\Delta e$
using the derivative $\frac{\partial \mathbf{M}}{\partial e} = \frac{\partial \mathbf{M}}{\partial \mathbf{I}} \cdot \frac{\partial \mathbf{I}}{\partial e}$.
These methods necessitate a fully differentiable chain, 
and require multiple updates to converge to the optimal solution.
Alternatively, function-fitting methods
derive the reverse function of imaging process, $\mathbf{I}^{-1}(\cdot)$.
By employing image synthesis techniques, 
these methods sample several seed exposures to generate corresponding images, 
thereby revealing the function $\mathbf{M}(\cdot)$ through these samples. 
Although more time-consuming, these methods can determine the optimal exposure in a single update.

\subsection{Our System Design}
In this work, we employ DRL to train agents for camera exposure control. 
Different from existing research~\cite{tomasi2021learned, lee2024learning} 
which directly output the dual control parameters from the agent network,
we split the exposure control problem into two distinct modules.
The first module, referred as $\mathbf{ModuleExpo}$, determines the optimal exposure $e$
for the coming image, formulated as
\begin{equation}
    e_{i+1} = \mathbf{ModuleExpo}(\mathbf{I}_{i-n+1:i}).
\end{equation}
$\mathbf{ModuleExpo}$ processes a sequence of $n$ historical images as input.
This module is offline trained by a DRL framework with our environment simulator.
The second module $\mathbf{ModuleAlloc}$ performs the attribute allocation role as
\begin{equation}
    t_{i+1}, g_{i+1} = \mathbf{ModuleAlloc}(\mathbf{I}_{i-n+1:i}, e_{i+1}).
\end{equation}
While not the primary focus of this work, 
this module is implemented using either the progressive strategy from our previous work~\cite{zhang2024image} 
or a motion-awareness method~\cite{han2023camera}.

Our framework design enables the training completely offline, 
avoiding online interaction with real-world scenarios.
This approach significantly enhances the efficiency of data collection and training, 
rendering it more feasible for practical applications. 
It enhances the efficiency of data collection and training and makes it more feasible in practical applications.
In subsequent sections of this paper, we will introduce our simulator implementation in Sect.\ref{sect_simulator}, 
detail the DRL design in Sect.\ref{sect_drl}, 
and demonstrate the effectiveness of our methods through experiments in Sect.~\ref{sect_exp}.

\section{Imaging Simulation Environment}\label{sect_simulator}
During the training phase, 
a simulated environment can efficiently mitigate the inefficiencies 
associated with direct agent interaction in real-world scenarios. 
While existing simulators like CARLA~\cite{dosovitskiy2017carla}
and Isaac Gym~\cite{liang2018gpu} are capable of supporting our tasks, 
yet they prove cumbersome or exhibit a notable discrepancy between the simulated imagery 
and actual real-world conditions.

To better align with our specific requirements, 
we have developed a lightweight simulator, 
which facilitates the replay of captured image sequences under varied exposure settings.
Additionally, we have incorporated a motion data augmentation module 
designed to increase the diversity of trajectories, 
utilizing a limited dataset of image sequences.

\subsection{Photometric Synthesis Process}
In the principles of imaging, 
exposure time $t$ is determined by camera shutter, 
whereas analog gain $g$ amplifies the electrical signal emanating from the sensor array.
The process of image synthesis can be viewed as another \emph{virtual amplification}; 
however, it cannot be directly applied to images 
due to the nonlinear transformation of electrical signals (or irradiance) into image intensities.
Photometric calibration~\cite{debevec2008recovering} is essential 
to elucidate the relationship between electrical signal and image intensity, 
termed as the camera response function~(CRF).
Typically, this relationship is characterized by an inverse mapping 
from image intensity to logarithmic irradiance, expressed as
\begin{equation}
    \mathbf{G}(\mathbf{I}) = \ln e + \ln \mathbf{E},
\end{equation}
where $\mathbf{E}$ denotes the irradiance associated with image $\mathbf{I}$, solely determined by the scene. 
Once the function $\mathbf{G}(\cdot)$ and its inverse $\mathbf{G}^{-1}(\cdot)$ are established, 
the synthesis process is
\begin{equation}
    \mathbf{I}_1 = \max(0, \min(255, \mathbf{G}^{-1}(\mathbf{G}(\mathbf{I}_0)-\ln e_0 + \ln e_1))).
\end{equation}

Inspired from Gamache~\textit{et al.}~\cite{gamache2023exposing}, 
we leverage bracketing image techniques to encompass a wide range of irradiances present within a scene.
Specifically, we employ a bracketing capture consisting of $5$ images~(Fig.~\ref{fig_simu_photometric})
with exposure times of $50$~us, $200$~us, $1$~ms, $5$~ms, and $20$~ms.
When an agent requires an image with a specific target exposure, 
we select the seed image whose exposure time is just below the target,
and then serves it as the basis for synthesizing, thus enabling effective interaction.

\subsection{Motion Augmentation}
We realize the necessity of incorporating a broad spectrum of motion dynamics within our training datasets.
However, the training trajectories remain invariant after collection. 
Additionally, employing a bracketing technique to widen the range of irradiance inherently 
diminishes the simulator's frame rate output. 
This reduction in frame rates indirectly affects the perceived speed of camera movement, 
thus introducing discrepancies between training and inference.

\begin{figure}[t]
    \centering
    \subfigure[Photometric synthesis process.
        \label{fig_simu_photometric}]{
        \centering
        \includegraphics[width=0.9\linewidth]{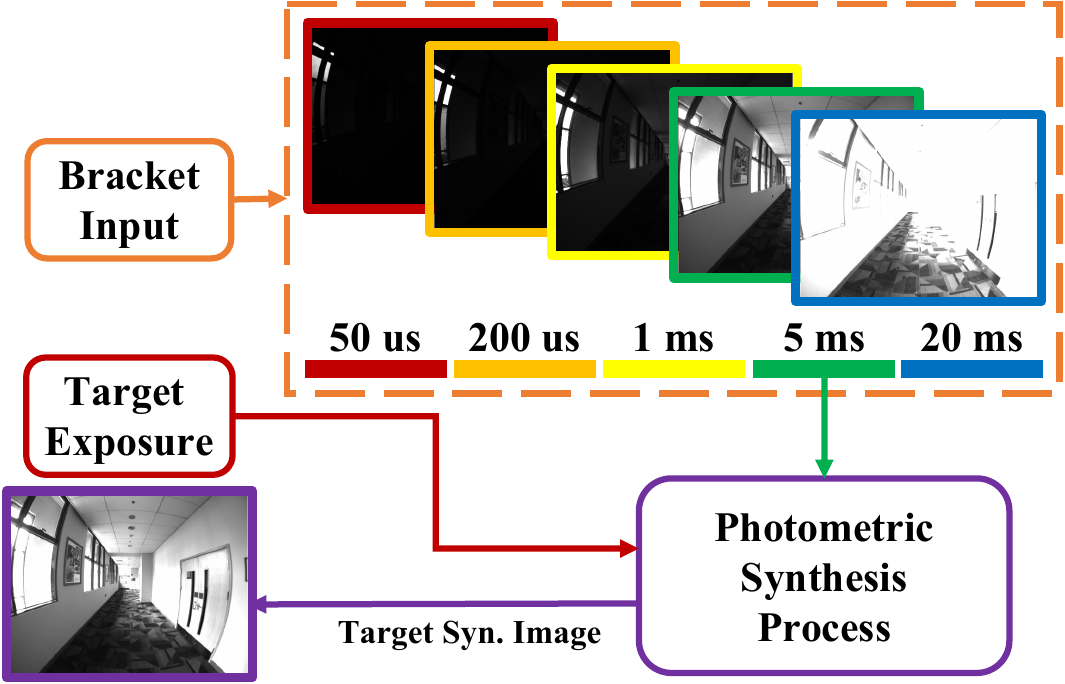}} \\
    \subfigure[Motion data augmentation.
        \label{fig_simu_motion}]{
        \centering
        \includegraphics[width=0.9\linewidth]{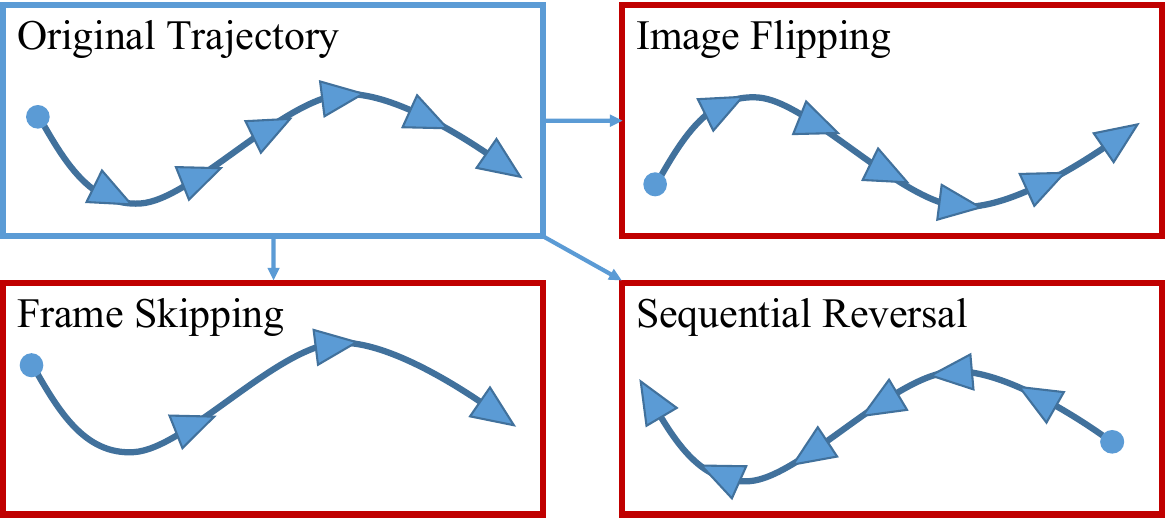}}
    \caption{Our image simulation environment for DRL training.
        The photometric synthesis module enables offline interaction between the agents and captured image sequences. 
        The motion data augmentation significantly enhances the diversity of available trajectories.}\label{fig_simulator}
\end{figure}

To address the limited variability in trajectories, 
we implement several motion augmentation strategies~(Fig.~\ref{fig_simu_motion}). 
These strategies are randomly combined and introduced during training.
\textbf{Image Flipping}:
Horizontal and vertical flips are applied to entire sequences 
to simulate diverse motion patterns and orientations.
\textbf{Speed Adjustment via Frame Skipping}:
Frame skipping is employed to artificially increase the playback speed 
by factors of $\times 1$, $\times 2$, $\times 3$, 
enabling the simulation of various motion velocities.
\textbf{Sequence Reversal}:
Sequences are reversed to model entirely different behaviors. 
For instance, a sequence transitioning from dark to light 
can be inverted into a shift from light to dark.

Diverging from~\cite{lee2024learning}, we abstain from cropping images during data augmentation.
By maintaining the integrity of the entire image frame, 
we facilitate the learning of lens distortion effects, 
enabling direct integration of these characteristics into the neural network.

\section{Reinforcement Learning Framework Design}\label{sect_drl}
The module $\mathbf{ModuleExpo}$ takes a segment of an image sequence as input 
and predicts the optimal exposure $e$ for subsequent scenes. 
We evaluated both Soft Actor-Critic (SAC) and Proximal Policy Optimization (PPO) for policy training. 
While both algorithms completed the training process successfully, 
PPO demonstrated a faster training speed but tended to converge to suboptimal policies. 
Consequently, we selected SAC as the preferred algorithm.

\subsection{State}
Our system processes $n=4$ historical images as inputs, 
which are resized to $84 \times 84$ pixels, 
yielding an input dimension of $4 \times 84 \times 84$. 
These inputs are first encoded by a CNN backbone consisting of three convolutional layers,
integrated into the Actor and the Critic respectively. 
Both Actor and the Critic employ two fully connected layers as the head, 
featuring a hidden layer with a dimension of $512$.

\subsection{Action}
Continuous action spaces have been employed by both previous work~\cite{tomasi2021learned, lee2024learning} due to 
their reduced need for update times compared to discrete spaces.
Tomasi~\textit{et al.} implemented an absolute action update scheme, 
in which the agent determines the absolute exposure time $t$ and gain $g$. 
We identify inefficiencies in this approach, 
as it needs referencing the exposure parameters of each image into the input 
to inform the network's output. 
Additionally, the requirement for the action output 
to span all potential exposures complicates convergence with limited training samples.
Lee \textit{et al.} opted for a relative action update scheme, 
where the agent outputs a relative difference in action values. 

We have developed a continuous-relative action scheme 
based on the concept of exposure compensation, 
quantified by the exposure value~(EV) as
\begin{equation}
    \text{EV} = \log_2{\frac{f^2}{t}},
\end{equation}
where $f$ represents the aperture value, and $t$ is the exposure time. 
An EV increment of $-1$ doubles the exposure time, 
whereas $+1$ halves it. 
We define the action value range for EV between $[-2, +2]$. 
allowing each update to potentially increase or decrease the current exposure by up to fourfold.

\subsection{Reward}
The design of the reward function is a pivotal component in reinforcement learning, 
as it directly influences the learning strategy of the agent. 
In this study, we investigated three distinct reward functions, 
each tailored to enhance convergence rates during training and to cultivate varying levels of agent intelligence.

\subsubsection{Statistical Reward}
The image statistical reward~($\mathcal{R}_{stat}$) aims to optimize image brightness and stability, 
defined as
\begin{equation}
    \mathcal{R}_{stat} = \mathcal{R}_{mean} - w_{flk} \cdot \mathcal{R}_{flk},
\end{equation}
where $\mathcal{R}_{mean}$ quantifies the average image intensity, 
and $\mathcal{R}_{flk}$ addresses flickering, 
as discussed by~\cite{lee2024learning}. 
The coefficient $w_{flk}=0.2$ balances the smoothness of exposure control.

\subsubsection{Feature Reward}
The feature reward~($\mathcal{R}_{feat}$) 
evaluates the effectiveness of feature point detection and tracking
\begin{equation}
    \mathcal{R}_{feat} = w_{detect} \cdot \mathcal{R}_{detect} + w_{match} \cdot \mathcal{R}_{match},
\end{equation}
where $\mathcal{R}_{detect}$ counts the detected feature points in the current frame, 
and $\mathcal{R}_{match}$ is the number of features matched between consecutive frames. 
The scaling coefficients $w_{detect} = 0.005$ and $w_{match} = 0.005$ ensure appropriate reward magnitudes, related to the feature numbers.
Feature detection employs the ORB descriptor~\cite{rublee2011orb}, 
and homography estimation refined with RANSAC filters outliers in feature matching.

\begin{table}
    \caption{Hyperparameters of SAC Training Phase}
    \begin{tabular}{llll}
        \toprule
        Parameter           & Value & Parameter                  & Value    \\
        \midrule
        Optimizer           & Adam  & Policy Distribution        & Gaussian \\
        Learning Rate       & 1e-4  & Buffer Size                & 50000    \\
        Alpha Learning Rate & 1e-4  & Warm-up Size               & 5000     \\
        Batch Size          & 256   & Discount Factor ($\gamma$) & 0.99     \\
        Episode Length      & 500   & Smooth Coefficent ($\tau$) & 0.005    \\
        \bottomrule
    \end{tabular}
    \label{table_sac_hyperparameters}
\end{table}

\subsubsection{Pose Error Reward}
The pose error reward~($\mathcal{R}_{pose}$) assesses the relative pose difference of VO systems 
by comparing it against ground truth data.
The pose derived from images is denoted as $\{\textbf{R}_{img} \in \mathbb{SO}(3), \textbf{t}_{img} \in \mathbb{R}^3\}$, 
encompassing both rotation and translation components.
This pose is calculated using the two-frame reconstruction from ORB-SLAM~\cite{campos2021orb}, 
which involves computing the fundamental and homography matrices from matched points 
and deriving transformations from both.
The pose with the minimal reprojection error is selected. 
A LiDAR-IMU odometry~(LIO) system~\cite{chen2022direct} 
outputs the ground truth trajectory, represented as $\{\textbf{R}_{gt} \in \mathbb{SO}(3), \textbf{t}_{gt} \in \mathbb{R}^3\}$.  
The reward formulation is
\begin{equation}
    \mathcal{R}_{pose} = -w_{rot} \cdot \mathcal{R}_{rot} - w_{trans} \cdot \mathcal{R}_{trans},
\end{equation}
where $w_{rot} = 10$ and $w_{trans} = 1$ serve as scaling factors.
The translation error is defined by
\begin{equation}
    \mathcal{R}_{trans} = || \frac{\textbf{t}_{img}}{|| \textbf{t}_{img} ||_2} - \frac{\textbf{t}_{gt}}{|| \textbf{t}_{gt} ||_2} ||_2,
\end{equation}
normalizing both translations before computing their Euclidean distance to account for scale discrepancies. 
The rotational error, $\mathcal{R}_{rot}$, is quantified in Lie algebra space~\cite{sola2018micro}
\begin{equation}
    \mathcal{R}_{rot} = || (\log(\textbf{R}_{img}^{-1} \cdot \textbf{R}_{gt}))^{\vee} ||_2,
\end{equation}
where $\log(\cdot)$ denotes the matrix logarithm operator, 
and $(\cdot)^{\vee}$ converts from Lie algebra to its vector space. 
$\mathcal{R}_{rot}$ also means the rotation angle in radians. 
To cap the reward, its maximum value is set to $1$, 
implying that rotation errors exceeding $57.3$ degrees do not further penalize the agent.

\begin{table*}[]
    \centering
    \caption{Performance comparison with ORB-SLAM3 using RPE~(MAX / RMSE).
        The number of relocalization is given in bracket if any.
        The marker '-' represents the failure cases, caused by system crash or RMSE RPE greater than $1.0$.}
    \setlength{\tabcolsep}{3.9pt}
    \begin{tabular}{cc|cccccccc}
        \toprule
        \multicolumn{2}{c}{Sequence}             & Built-in                      & Shim~\cite{shim2014auto}      & Zhang~\cite{zhang2017active}     & Kim~\cite{kim2020proactive}          & Zhang~\cite{zhang2024image}   & DRL-stat                      & DRL-feat                      & DRL-pose                         \\
        \midrule
        \multirow{2}{*}{Courtyard}   & $\times$1 & \textbf{1.42} / \textbf{0.29} & 2.02 / 0.31                   & 1.97 / 0.32                      &  2.23 / 0.31                         & 2.55 / 0.31                   & 2.17 / 0.31                   & 1.77 / 0.33                   & 2.04 / 0.35                      \\
                                     & $\times$2 & - (1)                         & 7.98 / 0.88                   & 6.78 / 0.71                      &  2.45 / 0.53                         & 1.64 / \textbf{0.49}          & \textbf{1.50} / \textbf{0.49} & 1.75 / 0.52                   & 2.38 / 0.58                      \\
        \midrule
        \multirow{2}{*}{Corridor}    & $\times$1 & 1.98 / 0.22                   & 2.24 / 0.24                   & 2.14 / 0.25                      &  2.26 / 0.25                         & 2.19 / 0.24                   & 1.96 / 0.23                   & \textbf{0.82} / \textbf{0.21} & 3.00 / 0.32                      \\
                                     & $\times$2 & 4.28 / 0.39 (1)               & 5.26 / 0.44 (1)               & 5.72 / 0.46 (1)                  &  6.12 / 0.56 (2)                     & 4.88 / 0.45 (1)               & 4.13 / 0.43 (1)               & \textbf{1.29} / \textbf{0.32} & 7.48 / 0.64 (2)                  \\
        \midrule
        \multirow{2}{*}{Parking}     & $\times$1 & 2.40 / 0.43                   & 3.14 / 0.43                   &  3.54 / 0.43                     &  - (1)                               & \textbf{1.44} / 0.42          & 2.34 / 0.42                   & 1.98 / \textbf{0.41}          & 2.69 / 0.51                      \\
                                     & $\times$2 & -  (1)                        & - (1)                         &  - (1)                           &  - (3)                               & - (2)                         & - (3)                         & \textbf{1.84} / \textbf{0.62} & -  (1)                           \\
        \midrule
        Switch                       & $\times$1 & 2.52 / 0.34 (1)               & 3.38 / 0.35 (1)               & 3.32 / 0.35 (1)                  &  2.02 / 0.30 (1)                     & 1.81 / 0.27 (1)               & 2.30 / 0.32 (1)               & 2.67 / 0.34 (1)               & \textbf{1.65} / \textbf{0.23}    \\
        \bottomrule
    \end{tabular}
    \label{tab_vo}
\end{table*}

\section{Experiment}\label{sect_exp}
\subsection{Implemetation Details}
\begin{figure}[t]
    \centering
    \includegraphics[width=0.9\linewidth]{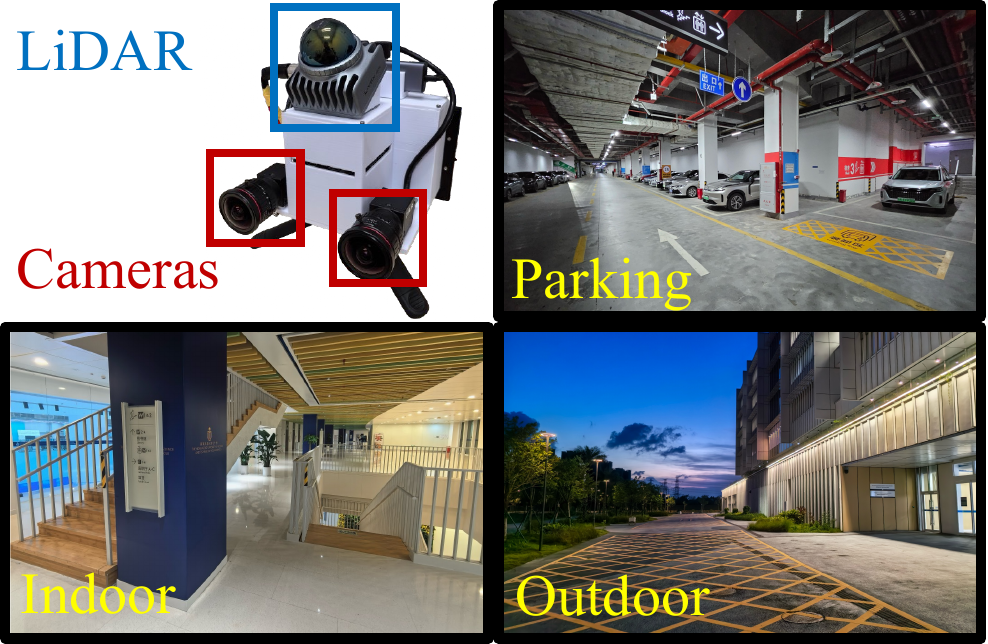}
    \caption{Our experimental platform and training data scenarios in campus.}\label{fig_setup}
\end{figure}

Our experimental setup depicted in Fig.~\ref{fig_setup},
comprises a portable PC equipped with an Intel i3-n305 CPU of low power consumption.
We selected two FLIR BFS 31S4C cameras, each fitted with a fisheye lens, 
to facilitate a wide field of view. 
To enhance image brightness in low-light conditions, we implemented a $4 \times 4$ binning pattern, 
yielding an output resolution of $512 \times 384$ pixels.
One camera utilizes bracketing patterns to capture data for training and evaluation, 
while the other employs a built-in exposure method to serve as a reference during evaluation.
A Livox Mid-360 LiDAR integrated with a built-in IMU 
preforms a LIO system~\cite{chen2022direct}, 
served as the ground truth trajectory for VO system evaluation.
Calibration of the intrinsic parameters of the cameras 
and the extrinsic parameters between the cameras and the IMU 
are conducted by using Kalibr~\cite{rehder2016extending}. 

The training datasets are captured on campus, as shown in Fig.~\ref{fig_setup}.
It comprises $3$ sequences, including indoor and outdoor campus environments, 
as well as the underground parking lot, totaling $29$ minutes and comprising $61653$ images.
The agents were trained using PyTorch on a PC with an Intel i9-13900k CPU and an Nvidia GeForce RTX 3080Ti GPU. 
The training process continued for $10000$ episodes with network updates occurring every $50$ frames.
The hyperparameters for SAC is shown in Table~\ref{table_sac_hyperparameters}.

\subsection{Experiment: Visual Odometry}
We evaluate the effectiveness and stability of different exposure control methods 
by processing image sequences through ORB-SLAM3~\cite{campos2021orb}. 
we generate exposure sequences using our simulator to ensure consistency in trajectories and image scenes, 
thereby mitigating errors due to variations in camera intrinsic and extrinsic parameters.
Given the limitations of LIO in providing accurate global positioning,
we adopt the relative pose error (RPE) as our evaluation metric, 
utilizing the \emph{evo} toolbox~\cite{rebecq2016evo}. 
We also record the number of failures and relocalizations across methods 
due to the challenging nature of the scenes.
For our evaluation, we captured four distinct sequences, 
each representing different lighting and environmental conditions:
\begin{itemize}
    \item \emph{Courtyard}: A typical VO environment with moderate lighting variations 
    caused by the occlusions from surrounding buildings.
    \item \emph{Corridor} and \emph{Parking}: Both sequences feature significant lighting variations 
    that challenge the exposure control methods, often leading to VO system failures.
    \item \emph{Switch}: This sequence tests the responsiveness of the system to unpredictable, rapid light changes.
\end{itemize}
To prevent training overfitting, these evaluation sequences were collected in locations different 
from those used for training.
For the sequences \emph{Courtyard}, \emph{Corridor}, and \emph{Parking}, 
we increased the challenge by doubling the playback speed, as described in Sect.~\ref{sect_simulator}.

Several methods are implemented as the baseline comparison.
\textbf{Built-in} is a feedback control method from FLIR cameras, 
aimed at optimizing average image brightness to a moderate value.
\textbf{Shim}~\cite{shim2014auto} and \textbf{Zhang}~\cite{zhang2017active} 
are two gradient-based methods target on optimizing image gradient magnitude and its variants. 
\textbf{Kim}~\cite{kim2020proactive} and \textbf{Zhang}~\cite{zhang2024image} are 
two function-fitting methods characterized by rapid parameter update speed.
Since we designed the three reward functions, 
we trained \textbf{DRL-stat}, \textbf{DRL-feat}, and \textbf{DRL-pose} for comparison.

Quantitative results are presented in Table~\ref{tab_vo}. 
In the \emph{Courtyard} sequence, 
performance across all methods is comparable with minimal variance. 
However, when operating at double speed, 
the \textbf{Built-in} was unable to adjust adequately to rapid motion, 
resulting in over-exposure and subsequent tracking failures.
In the sequences of \emph{Corridor} and \emph{Parking}, characterized by significant lighting variations, 
\textbf{DRL-feat} demonstrates superior performance. 
This advantage stems from its capability to swiftly predict motion-based events and rapidly adjust to scene parameters, 
thereby maintaining consistent interframe tracking even at double speed. 
In contrast, other methods fail to deliver rapid and stable control under conditions of swift motion coupled with dramatic lighting changes, 
resulting in degradation or failure of the VO systems.
In the \emph{Switch} sequence, characterized by unpredictable lighting changes of $4$ times,
all methods were interrupted by the most dramatic transition from dark to light, 
expect \textbf{DRL-pose}.
This is because \textbf{DRL-pose} prefers a lower exposure:
\textbf{DRL-pose} aims to directly minimize inter-frame pose errors.
This enhances the agent's preference for stable structural features over unstable textural features, which are plentiful but less reliable. 
Low exposure settings help obscure textural features and prevent blooming effects around structural features, 
which are typically delineated by contrasts between light and dark areas.

\begin{figure}[t]
    \centering
    \includegraphics[width=0.95\linewidth]{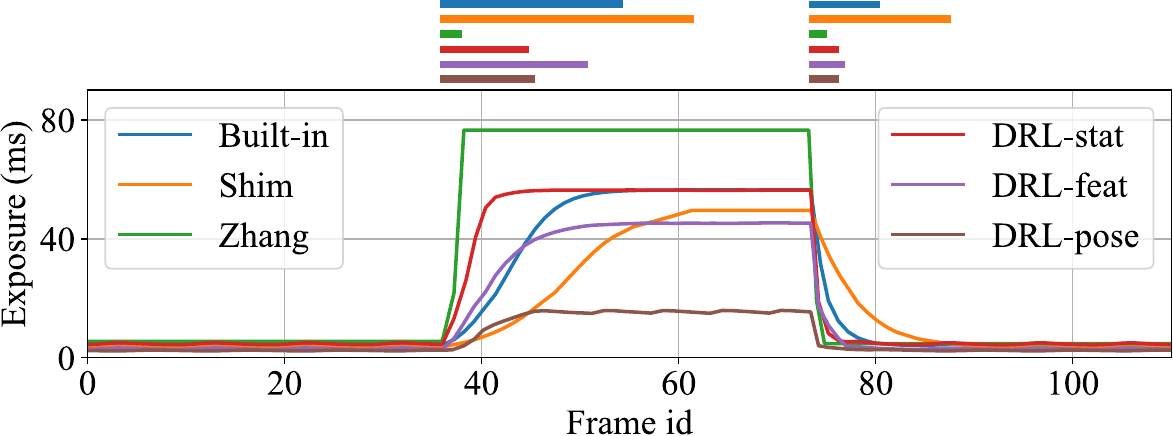}
    \caption{The experiment of the reaction speed with a switch case of light off then on.
    The method of \textbf{Zhang}~\cite{zhang2024image} performs the fastest because of its principle of one step control.
    All our DRL methods (\textbf{DRL-stat}, \textbf{DRL-feat}, and \textbf{DRL-pose}) respond faster than
    feedback control methods (\textbf{Built-in} and \textbf{Shim}).
    }\label{fig_exp_reaction}
\end{figure}

Additionally, we conducted a static experiment in the \emph{Switch} scene 
to compare the response speeds of different methods to abrupt unpredictable lighting changes. 
The response performance is illustrated in Fig.~\ref{fig_exp_reaction}. 
The function fitting method exhibited the quickest response, benefitting from its one-shot mechanism. 
Our DRL-based methods all respond faster than the traditional feedback control methods 
because their update hyperparameters are adaptively determined by the neural network.

\subsection{Experiment: Data Augmentation}
\begin{figure}[t]
    \centering
    \subfigure[Training process. \label{fig_exp_da_train}]{
        \centering
        \includegraphics[width=0.465\linewidth]{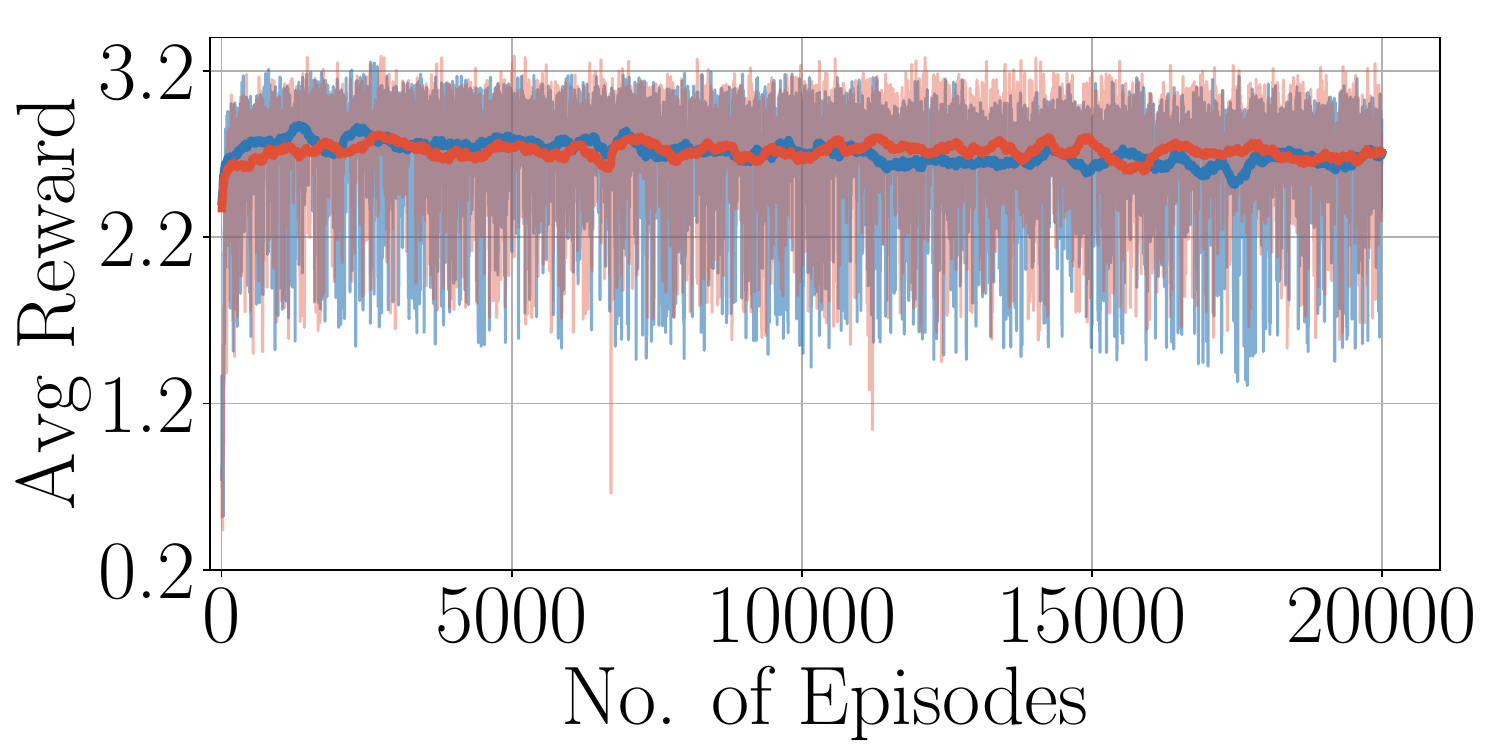}}
    \subfigure[Evaluation process. \label{fig_exp_da_eval}]{
        \centering
        \includegraphics[width=0.465\linewidth]{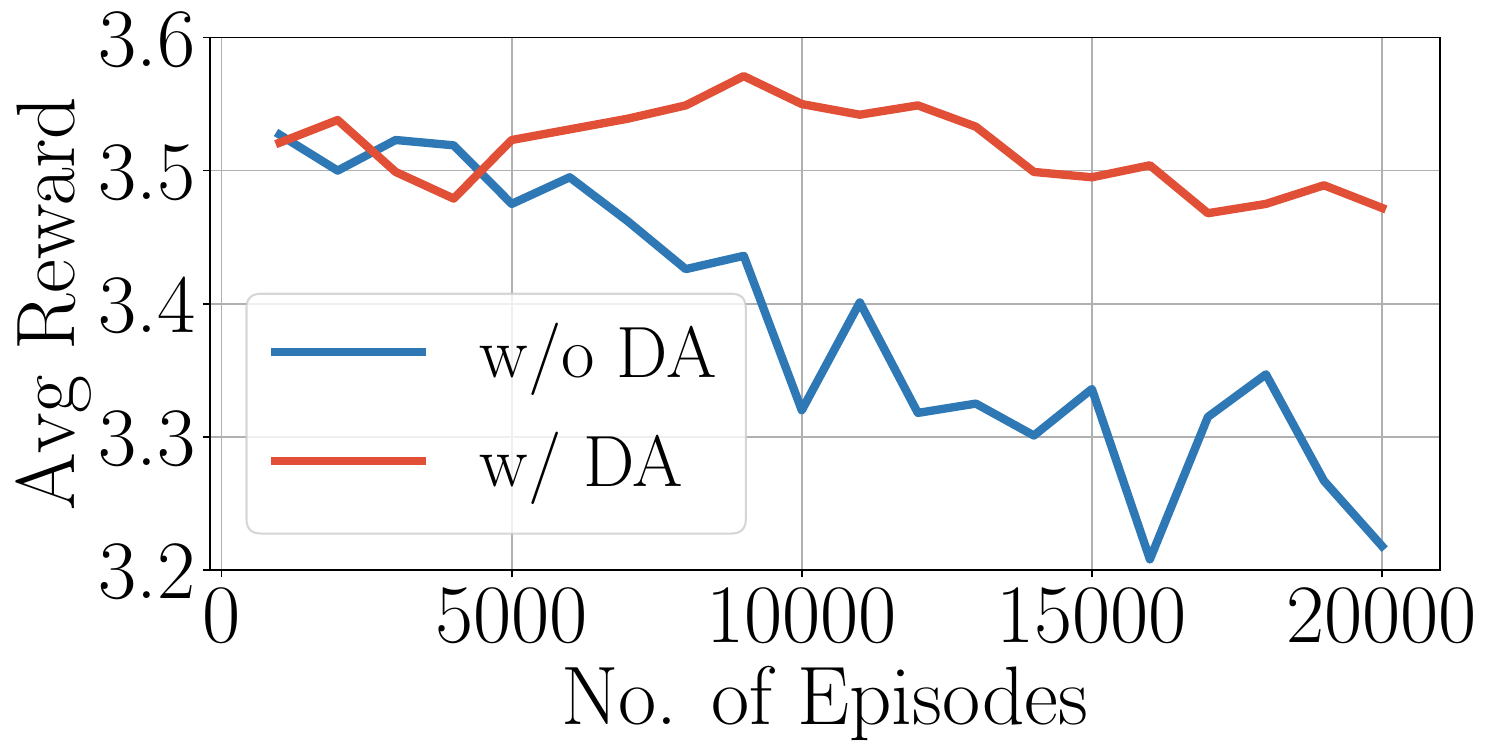}}
    \caption{The trend of average reward alongside the number of training episodes.
        The agent training results with and without our motion data augmentation.
        To clearly see the training trend, we introduced a median filter with a size of $100$.
        It can be seen that the training process without data augmentation (blue) has overfitting phenomenon, 
        and the average reward on the evaluation keeps decreasing.
    }\label{fig_exp_da}
\end{figure}

To assess the efficacy of our data augmentation module, 
we executed a controlled experiment wherein the \textbf{DRL-feat} agent 
was trained twice under identical hyperparameter settings. 
The only difference was the incorporation or exclusion of our motion augmentation module. 
Each agent underwent training across $20000$ episodes, 
with the average reward per episode meticulously recorded. 
For evaluation purposes, model parameters were archived at every $1000$ episode interval, 
and the average rewards were computed using the \emph{Corridor} sequence. 
The results of this experiment are illustrated in Fig.~\ref{fig_exp_da}.

After $2000$ training episodes, a plateau in average rewards was observed for both agents. 
However, convergence was not attained as indicated by the \emph{flickering} effect 
noted upon visual inspection of the \emph{Corridor} sequences from both agents. 
As training advanced, the agent equipped with the data augmentation module demonstrated a reduction in flickering 
and maintained consistent reward levels. 
Conversely, the agent devoid of the data augmentation experienced intensified flickering 
and a gradual decrease in average rewards, 
indicating overfitting to the training data.

\subsection{Experiment: Time Consumption}
Our training process does not necessitate any interaction with camera hardware or the real environment, 
rendering it significantly faster than previous approaches~\cite{tomasi2021learned, lee2024learning}. 
The training durations for our three agents—\textbf{DRL-stat}, \textbf{DRL-feat}, and \textbf{DRL-pose}—were 
$3$ hours $13$ minutes, $9$ hours $35$ minutes, and $10$ hours $33$ minutes, respectively, 
all trained for $10000$ episodes. 
The average training time per frame was recorded at $2.3$ ms, $6.9$ ms, and $7.6$ ms, respectively.
In comparison, the method developed by Lee~\cite{lee2024learning} with an online interaction platform, 
required an average of $100$ ms per frame. Consequently, our system demonstrates significantly higher training efficiency.

We further evaluated the inference cost of our DRL framework against several baseline methods. 
All baseline implementations utilized C++ with O2 optimization. 
In contrast, our DRL models were executed simply by using PyTorch on the CPU of a handheld device. 
The results of this comparison are documented in Table~\ref{table_time_comsumption}. 
The \textbf{Built-in} method exhibited the fastest performance due to its simpler computational operations. 
Our DRL agents significantly outpaced traditional methods 
and have the potential for further acceleration through software optimizations such as ONNX or TensorRT, 
or by leveraging dedicated camera hardware.

\section{Discussion}
\subsection{Reward Designs}\label{sect_diss_reward}
In this study, we designed three levels of reward functions, desiring to obtain different levels of intelligence for VO systems.
\textbf{DRL-stat} focuses on optimizing average brightness and performs comparably to traditional algorithms. 
However, the DRL framework offers faster and more precise control responses. 
Since it does not incorporate high-level feature information, 
it lacks the capability to understand the scene comprehensively and preemptively adapt to changes.
\textbf{DRL-feat} is designed to learn feature information with an objective to track as many feature points as possible. 
This capability enables it to comprehend motion trends and make predictions based on future static lighting conditions. 
However, it still suffers from overexposure during sudden, unpredictable light changes, 
though it recovers swiftly thanks to the DRL framework.
\textbf{DRL-pose} does not yield superior VO performance, 
primarily due to two factors. 
First, it relies on additional pose ground truth data from a LIO system, 
whose accuracy may not always be sufficient. 
Unlike the first two models, which solely depend on images, 
this introduces potential external errors. 
Second, it focuses on minimizing pose errors between \emph{adjacent} frames. 
Since these pose differences are minimal, triangulation errors can be significant. 
Although VO systems typically employ a keyframe technique to address this issue, 
attempts to integrate keyframe logic into the reward design did not result in model convergence. 
Keyframes led to sparse rewards, preventing the agent from receiving timely feedback.

\begin{table}[t]
    \centering
    \begin{threeparttable}
        \caption{Running Time Comparison}
        \begin{tabular}{ccccc}
            \toprule
            Method                & Built-in \tnote{1} & Shim  & Zhang & DRL  \\
            \midrule
            Time Comsumption (ms) & 0.39               & 11.77 & 15.09 & 1.58 \\
            \bottomrule
        \end{tabular}
        \begin{tablenotes}
            \item[1] Since \textbf{Built-in} is implemented by camera hardware, we reimplemented it for time cost measurement.
        \end{tablenotes}
        \label{table_time_comsumption}
    \end{threeparttable}
\end{table}

\subsection{Exposure Allocation}\label{sect_diss_alloc}
We also attempted to train a DRL agent for exposure allocation. 
This can constrain its time cost, since allocation methods using optical flow~\cite{han2023camera} is usually time-consuming. 
Consequently, the entire exposure control pipeline became differentiable, 
allowing for the joint optimization of agents after their initial separate training phases.

For training the exposure allocation module, 
we implemented an online training with another SAC, 
and also tried an offline training using a deep Q-Network~(DQN) 
based on parameter combinations from bracketing patterns. 
Both attempts employed our feature-level rewards. 
However, the agents were unable to develop stable or reasonable policies 
and failed to recognize blurring caused by camera motion.

The primary issue lies in the reward design, 
where a feature-level reward does not directly enhance the agent's perception of motion. 
We also explored image quality metrics related to motion blur. 
Given that each image lacks a reference ground truth, 
we were restricted to using non-reference image quality metrics, such as~\cite{crete2007blur}. 
However, the result shows that this kind of non-reference metrics is fragile for DRL training, 
ultimately leading to the unsuccessful training of the allocation agent.

\section{Conclusion}
In this work, we introduced deep reinforcement learning framework 
to train an exposure control agent for highly intelligent policies.
By designing a lightweight image simulator, 
we successfully offlined the training process, 
greatly reducing the training difficulty and time.
At the same time, we designed three different reward functions 
for the visual odometry tasks. 
The feature-level reward function enabled the agent to predict static lighting distribution and camera motion, 
and finally obtained advanced intelligence.
Experiments proved the efficiency of our method, 
and the performance improvement for subsequent VO systems.

\bibliographystyle{IEEEtran}
\bibliography{ref}{}

\begin{thebibliography}{10}
\providecommand{\url}[1]{#1}
\csname url@rmstyle\endcsname
\providecommand{\newblock}{\relax}
\providecommand{\bibinfo}[2]{#2}
\providecommand\BIBentrySTDinterwordspacing{\spaceskip=0pt\relax}
\providecommand\BIBentryALTinterwordstretchfactor{4}
\providecommand\BIBentryALTinterwordspacing{\spaceskip=\fontdimen2\font plus
\BIBentryALTinterwordstretchfactor\fontdimen3\font minus
  \fontdimen4\font\relax}
\providecommand\BIBforeignlanguage[2]{{%
\expandafter\ifx\csname l@#1\endcsname\relax
\typeout{** WARNING: IEEEtran.bst: No hyphenation pattern has been}%
\typeout{** loaded for the language `#1'. Using the pattern for}%
\typeout{** the default language instead.}%
\else
\language=\csname l@#1\endcsname
\fi
#2}}

\bibitem{shim2014auto}
I.~Shim, J.-Y. Lee, and I.~S. Kweon, ``Auto-adjusting camera exposure for
  outdoor robotics using gradient information,'' in \emph{2014 IEEE/RSJ
  International Conference on Intelligent Robots and Systems}.\hskip 1em plus
  0.5em minus 0.4em\relax IEEE, 2014, pp. 1011--1017.

\bibitem{zhang2017active}
Z.~Zhang, C.~Forster, and D.~Scaramuzza, ``Active exposure control for robust
  visual odometry in hdr environments,'' in \emph{2017 IEEE international
  conference on robotics and automation (ICRA)}.\hskip 1em plus 0.5em minus
  0.4em\relax IEEE, 2017, pp. 3894--3901.

\bibitem{han2023camera}
B.~Han, Y.~Lin, Y.~Dong, H.~Wang, T.~Zhang, and C.~Liang, ``Camera attributes
  control for visual odometry with motion blur awareness,'' \emph{IEEE/ASME
  Transactions on Mechatronics}, 2023.

\bibitem{kim2020proactive}
J.~Kim, Y.~Cho, and A.~Kim, ``Proactive camera attribute control using bayesian
  optimization for illumination-resilient visual navigation,'' \emph{IEEE
  Transactions on Robotics}, vol.~36, no.~4, pp. 1256--1271, 2020.

\bibitem{zhang2024image}
S.~Zhang, J.~He, B.~Xue, J.~Wu, P.~Yin, J.~Jiao, and M.~Liu, ``An image
  acquisition scheme for visual odometry based on image bracketing and online
  attribute control,'' in \emph{2024 IEEE International conference on robotics
  and automation (ICRA)}.\hskip 1em plus 0.5em minus 0.4em\relax IEEE, 2024.

\bibitem{tomasi2021learned}
J.~Tomasi, B.~Wagstaff, S.~L. Waslander, and J.~Kelly, ``Learned camera gain
  and exposure control for improved visual feature detection and matching,''
  \emph{IEEE Robotics and Automation Letters}, vol.~6, no.~2, pp. 2028--2035,
  2021.

\bibitem{lee2024learning}
K.~Lee, U.~Shin, and B.-U. Lee, ``Learning to control camera exposure via
  reinforcement learning,'' in \emph{Proceedings of the IEEE/CVF Conference on
  Computer Vision and Pattern Recognition}, 2024, pp. 2975--2983.

\bibitem{romero2023actor}
A.~Romero, Y.~Song, and D.~Scaramuzza, ``Actor-critic model predictive
  control,'' \emph{arXiv preprint arXiv:2306.09852}, 2023.

\bibitem{cai2022dqgat}
P.~Cai, H.~Wang, Y.~Sun, and M.~Liu, ``Dq-gat: Towards safe and efficient
  autonomous driving with deep q-learning and graph attention networks,''
  \emph{IEEE Transactions on Intelligent Transportation Systems}, pp. 1--1,
  2022.

\bibitem{jeong2021deep}
H.~Jeong, H.~Hassani, M.~Morari, D.~D. Lee, and G.~J. Pappas, ``Deep
  reinforcement learning for active target tracking,'' in \emph{2021 IEEE
  International Conference on Robotics and Automation (ICRA)}.\hskip 1em plus
  0.5em minus 0.4em\relax IEEE, 2021, pp. 1825--1831.

\bibitem{messikommer2024reinforcement}
N.~Messikommer, G.~Cioffi, M.~Gehrig, and D.~Scaramuzza, ``Reinforcement
  learning meets visual odometry,'' \emph{arXiv preprint arXiv:2407.15626},
  2024.

\bibitem{cheng2024pluto}
J.~Cheng, Y.~Chen, and Q.~Chen, ``Pluto: Pushing the limit of imitation
  learning-based planning for autonomous driving,'' \emph{arXiv preprint
  arXiv:2404.14327}, 2024.

\bibitem{xing2024bootstrapping}
J.~Xing, A.~Romero, L.~Bauersfeld, and D.~Scaramuzza, ``Bootstrapping
  reinforcement learning with imitation for vision-based agile flight,''
  \emph{arXiv preprint arXiv:2403.12203}, 2024.

\bibitem{dosovitskiy2017carla}
A.~Dosovitskiy, G.~Ros, F.~Codevilla, A.~Lopez, and V.~Koltun, ``Carla: An open
  urban driving simulator,'' in \emph{Conference on robot learning}.\hskip 1em
  plus 0.5em minus 0.4em\relax PMLR, 2017, pp. 1--16.

\bibitem{liang2018gpu}
J.~Liang, V.~Makoviychuk, A.~Handa, N.~Chentanez, M.~Macklin, and D.~Fox,
  ``Gpu-accelerated robotic simulation for distributed reinforcement
  learning,'' in \emph{Conference on Robot Learning}.\hskip 1em plus 0.5em
  minus 0.4em\relax PMLR, 2018, pp. 270--282.

\bibitem{debevec2008recovering}
P.~E. Debevec and J.~Malik, ``Recovering high dynamic range radiance maps from
  photographs,'' in \emph{ACM SIGGRAPH 2008 classes}, 2008, pp. 1--10.

\bibitem{gamache2023exposing}
O.~Gamache, J.-M. Fortin, M.~Boxan, F.~Pomerleau, and P.~Gigu{\`e}re,
  ``Exposing the unseen: Exposure time emulation for offline benchmarking of
  vision algorithms,'' \emph{arXiv preprint arXiv:2309.13139}, 2023.

\bibitem{rublee2011orb}
E.~Rublee, V.~Rabaud, K.~Konolige, and G.~Bradski, ``Orb: An efficient
  alternative to sift or surf,'' in \emph{2011 International conference on
  computer vision}.\hskip 1em plus 0.5em minus 0.4em\relax IEEE, 2011, pp.
  2564--2571.

\bibitem{campos2021orb}
C.~Campos, R.~Elvira, J.~J.~G. Rodr{\'\i}guez, J.~M. Montiel, and J.~D.
  Tard{\'o}s, ``Orb-slam3: An accurate open-source library for visual,
  visual--inertial, and multimap slam,'' \emph{IEEE Transactions on Robotics},
  vol.~37, no.~6, pp. 1874--1890, 2021.

\bibitem{chen2022direct}
K.~Chen, B.~T. Lopez, A.-a. Agha-mohammadi, and A.~Mehta, ``Direct lidar
  odometry: Fast localization with dense point clouds,'' \emph{IEEE Robotics
  and Automation Letters}, vol.~7, no.~2, pp. 2000--2007, 2022.

\bibitem{sola2018micro}
J.~Sola, J.~Deray, and D.~Atchuthan, ``A micro lie theory for state estimation
  in robotics,'' \emph{arXiv preprint arXiv:1812.01537}, 2018.

\bibitem{rehder2016extending}
J.~Rehder, J.~Nikolic, T.~Schneider, T.~Hinzmann, and R.~Siegwart, ``Extending
  kalibr: Calibrating the extrinsics of multiple imus and of individual axes,''
  in \emph{2016 IEEE International Conference on Robotics and Automation
  (ICRA)}.\hskip 1em plus 0.5em minus 0.4em\relax IEEE, 2016, pp. 4304--4311.

\bibitem{rebecq2016evo}
H.~Rebecq, T.~Horstsch{\"a}fer, G.~Gallego, and D.~Scaramuzza, ``Evo: A
  geometric approach to event-based 6-dof parallel tracking and mapping in real
  time,'' \emph{IEEE Robotics and Automation Letters}, vol.~2, no.~2, pp.
  593--600, 2016.

\bibitem{crete2007blur}
F.~Crete, T.~Dolmiere, P.~Ladret, and M.~Nicolas, ``The blur effect: perception
  and estimation with a new no-reference perceptual blur metric,'' in
  \emph{Human vision and electronic imaging XII}, vol. 6492.\hskip 1em plus
  0.5em minus 0.4em\relax SPIE, 2007, pp. 196--206.

\end{thebibliography}

\end{document}